\documentclass{ecai}
\usepackage{times}
\usepackage{graphicx}
\usepackage{latexsym}
\usepackage{amssymb}
\usepackage{amsmath}
\usepackage{xcolor}
\ecaisubmission   

\begin{document}

\title{Learning Global Pairwise Interactions with Bayesian Neural Networks}

\author{Tianyu Cui \institute{Department of Computer Science, Aalto University, Finland. Email: tianyu.cui@aalto.fi.} \and Pekka Marttinen* \institute{Department of Computer Science, Aalto University, Finland. Email: pekka.marttinen@aalto.fi. *Equal contribution.} \and Samuel Kaski* \institute{Department of Computer Science, Aalto University, Finland. Email: samuel.kaski@aalto.fi. *Equal contribution.} }

\maketitle
\bibliographystyle{ecai}

\begin{abstract}
Estimating global pairwise interaction effects, i.e., the difference between the joint effect and the sum of marginal effects of two input features, with uncertainty properly quantified, is centrally important in science applications. We propose a non-parametric probabilistic method for detecting interaction effects of unknown form. First, the relationship between the features and the output is modelled using a Bayesian neural network, capable of representing complex interactions and principled uncertainty. Second, interaction effects and their uncertainty are estimated from the trained model. For the second step, we propose an intuitive global interaction measure: Bayesian Group Expected Hessian (GEH), which aggregates information of local interactions as captured by the Hessian. GEH provides a natural trade-off between type I and type II error and, moreover, comes with theoretical guarantees ensuring that the estimated interaction effects and their uncertainty can be improved by training a more accurate BNN. The method empirically outperforms available non-probabilistic alternatives on simulated and real-world data. Finally, we demonstrate its ability to detect interpretable interactions between higher-level features (at deeper layers of the neural network).
\end{abstract}

\section{Introduction}
\label{sec: intro}
%

Estimating interactions between variables, and the uncertainty of the interactions, is a challenge common to many data science tasks, and ubiquitously needed in scientific discovery. For instance, gene-gene, gene-environment, gene-drug and gene-disease interactions are key elements in explaining genetic mechanisms, diseases and drug effects in genetics and health applications. Having the right interactions in a model also makes it more understandable and interpretable. At the simplest this could mean that the output $y$ depends on inputs $x_1$ and $x_2$ through
\begin{equation}
    y=\beta_1 x_1 + \beta_2 x_2 + \beta_{12}x_1 x_2 + e,
    \label{eq:simple_interaction}
\end{equation}
where $\beta_1$ and $\beta_2$ represent the \textit{main effects}, $\beta_{12}$ is the strength of the \textit{interaction}, and $e$ is noise. In this example, the shape of the interaction is known (multiplicative), but this is not true in general. Estimating the uncertainty is equally important, to assess the statistical significance of the detected interactions. Traditional methods include two general approaches: 1) conducting tests for each combination of features, such as  ANOVA based methods \cite{fisher1936statistical,wonnacott1990introductory} and information gains \cite{jiang2015learning,zeng2016discovering}. These method usually require a polynomial number of tests, and lack statistical power due to multiple testing; 2) interactions are first learned by 'white-box' machine learning models, and then recover the interaction effects from the trained model. For example, Lasso based methods \cite{bien2013lasso,kong2017interaction,lim2015learning}, and Gaussian processes \cite{agrawal2019kernel}. But all possible interactions have to be pre-specified in this approach, which restricts the type of interactions that can be considered to, for example, multiplicative as in Eq.\ref{eq:simple_interaction}.

 In this paper we extend the second approach, and use the representation learning power of a neural network to model interactions without any specified form directly from the data. Intuitively, we first train a neural network on the data, then find the encoded interactions by interpreting the trained model. However, the currently available algorithms that both are \textit{interpretable} and aim to recover all kinds of interactions \cite{greenside2018discovering,lundberg2018consistent,sorokina2008detecting,tsang2017detecting} neglect uncertainty, and thus it is not possible to conduct hypothesis tests and control type I and II errors, which is not acceptable in critical fields such as healthcare. Another limitation is that most algorithms mentioned before aim to find \textit{local} interactions (dependent on the location in the data space) instead of \textit{global} interactions (independent of the location), which is not suitable for most science discovery applications.
 
 In this work, we propose Bayesian Group Expected Hessian (GEH) to estimate global interactions by aggregating local interactions from a trained Bayesian neural network. The posterior distribution of GEH represents the uncertainty of the interaction measure, and it can be seen as a non-parametric analogy to the posterior distribution of $|\beta_{12}|$ in Eq.\ref{eq:simple_interaction} which could be learned using a Bayesian linear regression model. Two theoretical significant advantages of Bayesian GEH are: 1. it provides a natural trade-off between type I and type II error by tuning the number of groups; 2. estimated interaction effects can be improved by training a more accurate and better calibrated Bayesian NN, which is not guaranteed by existing methods that use NNs to learn interactions.

\section{Related Work}
\label{sec: literature}
Several approaches, such as neural networks and ensemble trees, have been proposed for detecting interactions with unknown forms.  \textit{Neural Interaction Detection} (NID) \cite{tsang2017detecting}  was proposed to learn interactions by inspecting the weights of a neural network: if some features are connected to the same node in the first hidden layer with large weight and if the path(s) from the node to the output also have large weights, then an interaction between the features is detected. An interpretable neural network architecture \cite{tsang2018neural} that contains interaction information has also been proposed based on this idea. These methods can detect higher-order interactions of unrestricted forms, but they can only be applied to fully connected vanilla multi-layer perceptrons, not to convolutional neural networks for instance, which limits their usage in applications such as computer vision. \textit{Deep Feature Interaction Map} \cite{greenside2018discovering} is similar to our work in that it uses the Hessian, but it is designed for discrete features only, such as DNA sequences. SHAP interaction score \cite{lundberg2018consistent} extends the SHAP value of game theory to detect interactions, and it is efficiently implemented in ensemble tree methods. Additive Groves \cite{sorokina2008detecting} is another tree-based method, which compares the difference in performance of two regression trees, one with all interactions, and the other with the interaction of interest removed. However, none of these methods estimates the uncertainty of the learned interactions, and thus is unable to control two types of error.

Bayesian Neural Networks (BNN) \cite{mackay1992practical,neal2012bayesian} have been studied since the 90s, but they have not been generally applicable to large datasets until scalable variational inference algorithms \cite{hoffman2013stochastic,kingma2013auto} were proposed. Recent works \cite{gal2015dropout,kingma2015variational} show that the standard dropout training in NNs can be interpreted as a form of variational inference, and that the prediction uncertainty can be obtained by simply using dropout during testing. This approach can be applied to most neural networks, such as RNNs \cite{gal2016theoretically}, and leads to sparse neural networks \cite{molchanov2017variational}. By optimizing dropout probabilities, calibration of uncertainty can be improved \cite{gal2017concrete}. In this work, we use the concrete dropout from \cite{gal2017concrete}, and extend the model by including a separate component for main effects to better capture interactions with uncertainty.
\section{Modeling Interactions and their Uncertainty}
\label{sec: modeling}
To learn interactions and their uncertainty, we first train a Bayesian neural network (BNN), which both models a rich family of functions that automatically incorporates all kinds of interactions, and captures uncertainty.
\subsection{Practical Bayesian Neural Network}
Bayesian neural networks are defined by placing a prior distribution on the weights $\mathbf{W}$ of a neural network. Then, instead of finding a point estimate of the weights by minimizing a cost function, a posterior distribution of the weights is calculated conditionally on the data. Let $g^{\mathbf{w}}(\mathbf{x})$ denote the output of a BNN and $p(\mathbf{y}|g^{\mathbf{w}}(\mathbf{x}))$ the likelihood. Then, given a dataset $\mathbf{X}=\{\mathbf{x}^{(1)},\ldots,\mathbf{x}^{(N)}\}$, $\mathbf{Y}=\{\mathbf{y}^{(1)},\ldots,\mathbf{y}^{(N)}\}$, training a BNN is equivalent to learning the posterior distribution $p(\mathbf{W}|\mathbf{X},\mathbf{Y})$. Variational inference can be used to approximate the intractable $p(\mathbf{W}|\mathbf{X},\mathbf{Y})$ with a simpler distribution, $q_{\theta}(\mathbf{W})$, by minimizing the $\text{KL}(q_{\theta}(\mathbf{W})||p(\mathbf{W}|\mathbf{X},\mathbf{Y}))$. This is equivalent to minimizing the negative ELBO
\begin{equation}
\begin{split}
    \mathcal{L}(\theta)=&\int -q_{\theta}(\mathbf{W})\log p(\mathbf{Y}|g^{\mathbf{w}}(\mathbf{X}))\text{d}\mathbf{W} + \text{KL}(q_{\theta}(\mathbf{W})||p(\mathbf{W})).
    \label{eq:elbo}
\end{split}
\end{equation}

According to recent research on dropout variational inference \cite{gal2015dropout}, a practical Bayesian neural network for a wide variety of architectures can be obtained by simply training a neural network with dropout (\textit{MC dropout}), and interpreting this as being equivalent to maximizing Eq.\ref{eq:elbo}. \textit{Concrete dropout} \cite{gal2017concrete} extends this by learning a different Bernoulli dropout probability for each layer to obtain better-calibrated uncertainty. In this case the variational distribution is defined as $q_{\mathbf{p}, \mathbf{M}}(\mathbf{W}) = \prod_{l=1}^{L}\prod_{k = 1}^{K_{l}}\mathbf{m}_{l,k}\text{Bernoulli}(1-p_{l})$, where $p_{l}$ is the dropout probability for layer $l$, and $\mathbf{m}_{l,k}$ is a vector of outgoing weights from node $k$ in layer $l$, and the prior $p(\mathbf{W})$ is a Gaussian distribution with a fixed length-scale. Eq.\ref{eq:elbo} can be optimized w.r.t. the variational parameters $\theta = \{\mathbf{p}, \mathbf{M}\}$ by using the re-parametrization trick and concrete relaxation \cite{gal2017concrete}. In this paper, we adopt concrete dropout BNNs to learn interactions and corresponding uncertainty. However, instead of learning a single dropout probability for \textit{each layer}, we learn a separate dropout probability for \textit{each node}. This not only improves the calibration of uncertainty, but the learned dropout probabilities also provide information about importance of features \cite{chang2017dropout}. When the dimension is high, we can reduce the computation by detecting interactions only among important features (with low dropout probabilities). Note that other BNNs can also be used to model interactions, and we consider an alternative BNN formulation using the mean-field algorithm in Appendix 4.

\subsubsection{Separating Interactions from Main Effects}
Real world data often contain both main effects and interactions, where a main effect represents the influence of one of the features on the target alone, irrespective of the other features. Interaction is then defined as the joint effect of multiple features that cannot be explained by the main effects \cite{winer1962statistical}.

In practice it is beneficial for deterministic NNs to model the main effects separately from the interactions  \cite{cheng2016wide,crane2018machine,tsang2017detecting}, and we extend this idea to Bayesian NNs. We will use linear regression, $\mathbf{y}^{m} = \mathbf{\beta}^{T}\mathbf{x}$, for the main effects, and a BNN, $\mathbf{y}^{in} = g^{\mathbf{W}}(\mathbf{x})$, to capture the interactions. A univariate NN could be applied to each feature to model non-linear main effects, but investigating this further is beyond the scope of the present work. A prediction from this \textit{hybrid} model is the sum of the two components $\hat{\mathbf{y}}=\mathbf{y}^{m}+\mathbf{y}^{in}$, such that the objective function in Eq.\ref{eq:elbo} can be rewritten as
\begin{equation}
\begin{split}
    \mathcal{L}(\theta, \beta)=&\int -q_{\theta}(\mathbf{W})\log p(\mathbf{Y}|\mathbf{\beta}^{T}\mathbf{X} + g^{\mathbf{w}}(\mathbf{X}))\text{d}\mathbf{W} \\
    &+ \text{KL}(q_{\theta}(\mathbf{W})||p(\mathbf{W})).
    \label{eq:elbo_main}
\end{split}
\end{equation}
If the uncertainty of main effects is of interest, a prior may be placed also on $\beta$. However, the prior on $\beta$ should impose weaker regularization than the prior on $\mathbf{W}$, otherwise the main effect would not be separated properly. The motivation for this model structure stems from the fact that in real-world data the main effects usually dominate interactions, such as genomics data. A BNN $g^{\mathbf{W}}(\mathbf{x})$ with a small capacity would likely capture the main effects only, unless they are modeled separately. By separating the main effects we can use a much smaller BNN to learn the interactions.

\section{Detecting Interactions}\label{sec: detecting}
After a BNN $g^{\mathbf{W}}(\mathbf{x})$ has been learned from data, interactions must be detected from the BNN. Previously, the gradient of $g^{\mathbf{W}}(\mathbf{x})$ has been used to detect main effects from a neural network \cite{bach2015pixel,hechtlinger2016interpretation,shrikumar2017learning,sundararajan2017axiomatic}. We extend these works and formulate Bayesian Group Expected Hessian (GEH), which now quantifies \textit{global} interactions with uncertainty, and can be easily calculated for any given BNN.

\subsection{Local Interaction Aggregation}
One way to measure interactions is to use the \textit{Hessian} of $g^{\mathbf{W}}(\mathbf{x})$ w.r.t. the input. For the simple multiplicative interaction in Eq.\ref{eq:simple_interaction} this recovers the coefficient $\beta_{12}$. For non-multiplicative interactions the Hessian is not constant and represents interaction only at the point at which it was calculated, making it a \textit{local interaction} measure. To estimate \textit{global interaction} effects (interaction regardless of the evaluation point), an intuitive way is to aggregate local effects into one global effect \cite{erion2019learning, van2019global}, and we propose below three ways of aggregating local interactions.
\subsubsection{Expected Absolute Hessian (EAH)} EAH aggregates point-wise Hessian by calculating the expectation of its absolute value. To estimate an interaction between features $x_i$ and $x_j$, we define $\text{EAH}^{i,j}_{g}$, as
\begin{equation}
    \text{EAH}^{i,j}_{g}(\mathbf{W}) =\mathbb{E}_{p(\mathbf{x})}\Big[\Big|\frac{\partial^2 g^{\mathbf{W}}(\mathbf{x})}{\partial x_{i}\partial x_{j}}\Big|\Big],
    \label{eq: EAH}
\end{equation}
where $p(\mathbf{x})$ is the empirical distribution of $\mathbf{x}$. Similar idea has been widely used in aggregating local marginal effects of features in \cite{lundberg2018consistent,van2019global}. EAH has the \textbf{lowest FNR} (False Negative Rate or type II error) of the three methods, because as long as there is some region in the domain of $\mathbf{x}$ ($\text{dom}(\mathbf{x})$) where the corresponding Hessian is non-zero, $\text{EAH}^{i,j}_{g}$ will also be non-zero. However, EAH has the \textbf{highest FPR} (False Positive Rate or type I error), since even when $x_i$ and $x_j$ do not interact with each other in the data generating process, the input Hessian of $g^{\mathbf{W}}(\mathbf{x})$ between $x_i$ and $x_j$ will not be exactly $0$ due to noise inherent in the data set and inevitable modeling error. This noisy effect will then be also aggregated into the global interaction effect according to Eq.\ref{eq: EAH}, and thus false interactions may be detected.
\subsubsection{Absolute Expected Hessian (AEH)} AEH is defined as the absolute value of expected Hessian, given by
\begin{equation}
    \text{AEH}^{i,j}_{g}(\mathbf{W}) =\Big|\mathbb{E}_{p(\mathbf{x})}\Big[\frac{\partial^2 g^{\mathbf{W}}(\mathbf{x})}{\partial x_{i}\partial x_{j}}\Big]\Big|.
    \label{eq: AEH}
\end{equation}
In contrast with EAH, AEH has the \textbf{lowest FPR} but the \textbf{highest FNR}. To understand this, assume that noise is distributed with zero mean independently of the location in $\text{dom}(\mathbf{x})$. Then, the noise will cancel when we take the expectation over $\text{dom}(\mathbf{x})$. Consequently, FPR will be low. On the other hand, AEH may also cancel out some true interactions, if some subregions of $\text{dom}(\mathbf{x})$ have positive and some subregions negative interaction effects, leading to a high FNR. AEH can be regarded as an extension of an expected gradient method \cite{erion2019learning}, where marginal feature effects are aggregated.

\subsubsection{Group Expected Hessian (GEH)}\label{sec:geh} One intuitive way to trade-off between EAH and AEH, and thus between FPR and FNR, is by clustering $\text{dom}(\mathbf{x})$ into \textit{subregions}, calculating AEH for each subregion, and then computing their weighted average. Based on this idea, we propose Group Expected Hessian with $M$ groups, $\text{M-GEH}^{i,j}_{g}$ as
\begin{equation}
\begin{split}
    \text{M-GEH}&^{i,j}_{g}(\mathbf{W}) = \sum_{m=1}^{M}\frac{|A_{m}|}{\sum_{k=1}^{M}|A_{k}|}\Big|\mathbb{E}_{p(\mathbf{x}|\mathbf{x}\in A_{m})}\Big[\frac{\partial^2 g^{\mathbf{W}}(\mathbf{x})}{\partial x_{i}\partial x_{j}}\Big]\Big|.
    \label{eq: GEH}
\end{split}
\end{equation}
where $|A_m|$ is the size of the $m$th subregion $A_m$, and $\bigcup_{m=1}^{M}A_m=\text{dom}(\mathbf{x})$.
The benefit of GEH is that, by choosing the subregions $A_m$ properly, it has the potential to aggregate only interactions while canceling out the noise. To see this, assume that the noise distribution is independent of location in $\text{dom}(\mathbf{x})$, but interaction effects are similar for close-by points in $\text{dom}(\mathbf{x})$. Then, a partition can be estimated by a clustering algorithm, such as k-means, with the number of clusters equal to $M$. Consequently, datapoints within $A_m$ are close to each other, and are expected to have interaction effects mostly of the same sign, which Eq.\ref{eq: GEH} will aggregate similarly to Eq.\ref{eq: EAH}. On the other hand, Eq.\ref{eq: GEH} will act like Eq.\ref{eq: AEH} canceling out the noise when integrating over the subregion. Similar ideas have been proposed to denoise and compress signals by vector quantization \cite{ayanoglu1990optimal,panchapakesan1998combined} in signal processing, where using less clusters can reduce more noise in the signal but also has a higher distortion error. When $M=1$, $\text{1-GEH}$ reduces to AEH, and when $M=N$, $\text{N-GEH}$ becomes EAH, where $N$ is the number of data. FNR can be reduced by increasing $M$ with the cost of increasing FPR. Another benefit of GEH is that we can immediately investigate the interaction effects in each subregion. Because the sign of a complex interaction may be different in different parts of the input space, investigating the subregional interactions can lead to interesting further insights about the data. We leave this for future work.

In Eq.\ref{eq: GEH}, $\mathbf{W}$ is the weight in a BNN, i.e. a random variable. Therefore, $\text{M-GEH}^{i,j}_{g}(\mathbf{W})$ is also a random variable whose distribution follows from the posterior distribution $q_{\theta}(\mathbf{W})$ of $\mathbf{W}$, thus we call it \textit{Bayesian Group Expected Hessian}. Unbiased estimators for the mean $\hat{m}^{i,j}$ and variance $\hat{v}^{i,j}$ of $\text{M-GEH}^{i,j}_{g}(\mathbf{W})$ can be obtained through Monte Carlo (MC) integration with $K$ samples. For concrete dropout BNN, MC integration only requires $O(KD)$ backward-passes with dropout, where $D$ is the number of features (see Appendix 1 for complexity analysis and a detailed algorithm).

To calculate the FPR, FNR, and ROC curves in our experiments, we determine the significance of an interaction between features $i$ and $j$ as follows. We assume $\text{M-GEH}^{i,j}_{g}(\mathbf{W})$ to be approximately Gaussian, which follows from the CLT for large $K$. Consequently, a 95\% credible interval is given by $(\hat{m}^{i,j}-2\sqrt{\hat{v}^{i,j}}, \hat{m}^{i,j}+2\sqrt{\hat{v}^{i,j}})$. The interaction is considered significant iff the 95\% CI does not contain 0. When comparing with deterministic alternatives which ignore uncertainty, we simply use the mean $\hat{m}^{i,j}$ of $\text{M-GEH}^{i,j}_{g}(\mathbf{W})$ to represent the interaction effect.

\begin{figure*}[ht]
    \centering
    \includegraphics[width=0.8\linewidth]{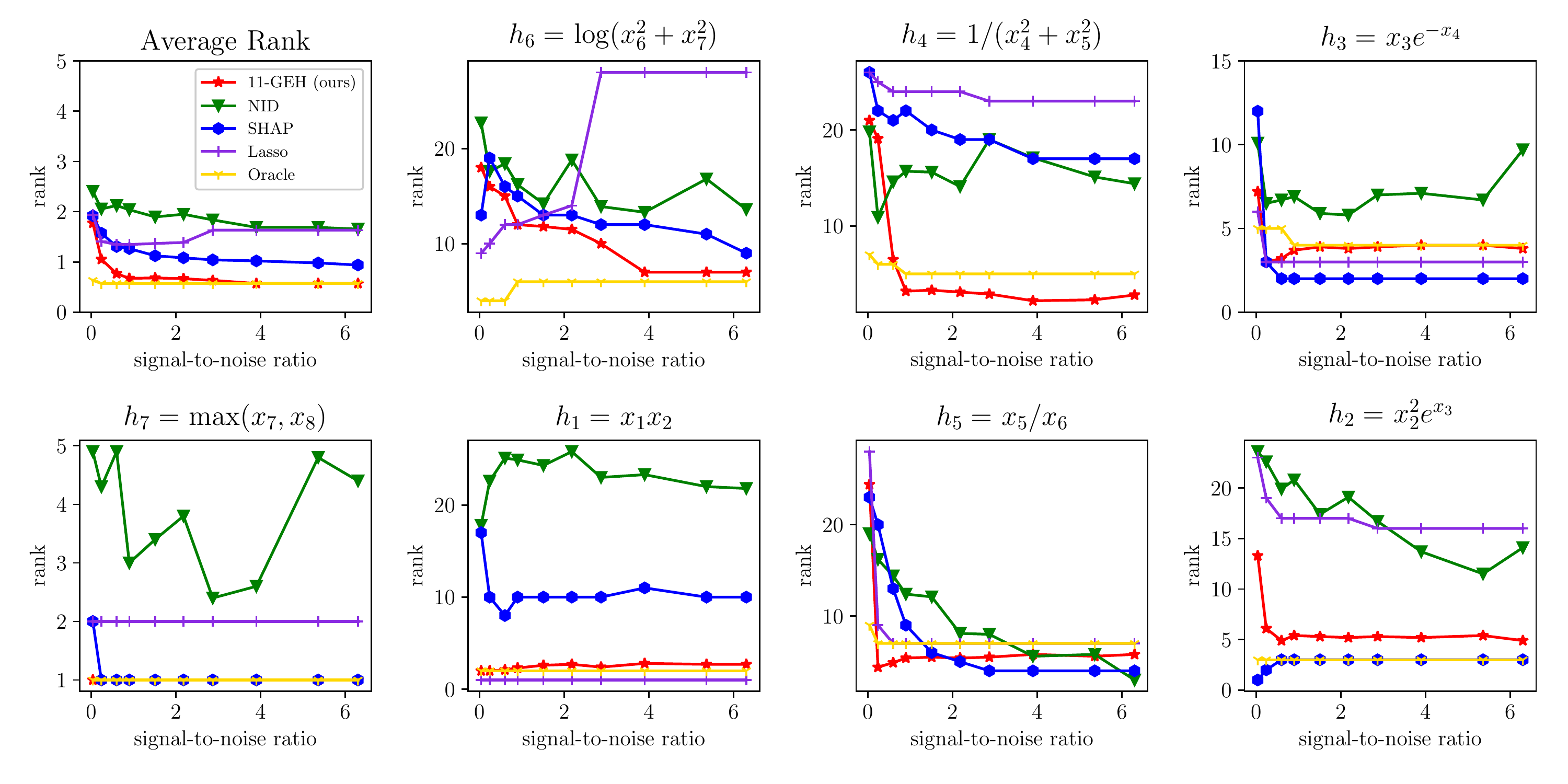}
    \caption{The ranks of true interactions as a function of S/N for different methods. The 11-GEH method detects the ground-truth interactions clearly better than the others (top left panel), with performance very close to the Oracle. The rest of the panels show the ranks assigned to each of the seven true interactions.}
    \label{fig: simulated data comparision}
\end{figure*}
\subsection{Accuracy Improvement Properties}
The error when learning interactions from data can be caused both by modeling error and error in detecting interactions from the model. Interaction detection algorithms, such as NID, intertwine these two errors, thus training a better NN does not guarantee learning of more accurate interaction effects. This makes it difficult to improve the detected interactions \cite{sundararajan2017axiomatic}. Suppose $f(\cdot)$ is the true data generating mechanism and $g(\cdot)$ is the learned model. If we further assume that $f(\cdot)$ and $g(\cdot)$ are Lipschitz functions, we can prove M-GEH has the following two properties (proofs are given in the Appendix 2):

\textbf{Property 1}  The estimation error in interaction measures, $\mathcal{L}_e = \sum_{i,j}\lvert \text{M-GEH}_{g}^{i,j}-\text{M-GEH}_{f}^{i,j}\rvert$, is linearly upper bounded by the prediction error $\epsilon$ of $g(\cdot)$.

\textbf{Property 2} When $g^{\mathbf{W}}(\cdot)$ is a probabilistic model (e.g. BNN), we can make the uncertainty of the Bayesian GEH arbitrarily well-calibrated by improving the calibration of the distribution of predictions from $g^{\mathbf{W}}(\cdot)$.

Intuitively, these two properties relate the accuracy of predictions to the accuracy of interpretations (interaction effects in this case). This is needed since testing whether the estimated interactions are accurate or not is almost impossible in practice, because of the lack of a ground truth. Here we use calibration to represent the 'accuracy' of the estimated uncertainty \cite{gneiting2007probabilistic}. In detail, the calibration of a distribution is defined in terms of its 'frequentist properties'. For instance, a 95\% credible interval is well-calibrated if it contains the true value 95\% of times. Property 1 guarantees that by training a better model with a smaller $\epsilon$, $\mathcal{L}_e$ will have a tighter upper bound, which improves the learning of interactions. Property 2 provides a way to improve the uncertainty of Bayesian GEH, i.e., by using a better calibrated BNN (more discussion and empirical evidence can be found in Appendix 4). 

\subsection{Determining the Number of Clusters} By increasing the number of clusters, GEH can be more flexible to detect more complex interactions, but can also lead to a higher FPR. An ideal number of cluster $M$ is the smallest number of clusters that can capture rich enough interactions for a specific problem, which means the detected interactions should not change significantly by further increasing $M$. Inspired by Spearman's Rho distance \cite{ibrahim2019global}, we propose rank weighted distance to compare two attribution vectors (interaction effect vectors) corresponding to consecutive numbers of clusters:
\begin{equation}
\begin{split}
    \Delta_{M}^{2} =\sum_{i=1}^{L}(w_{M}(i)-w_{M-1}(i))^{2}(\pi_{M}(i)-\pi_{M-1}(i))^{2}.
    \label{eq: rank weighted distance}
\end{split}
\end{equation}
Here, $w_{M}(i)$ is the $i$th interaction effect with $M$ ($M\geq2$) clusters, $\pi_{M}(i)$ is the rank of $w_{M}(i)$ among the interactions, and $L$ is the number of interactions. We ignore the tiny changes of interaction effect vectors that do not result in a change of ranks to emphasize the importance of finding the top ranked interactions, which also leads to a much smoother $ \Delta_{M}^{2}$. In Eq.\ref{eq: rank weighted distance}, the contribution to $\Delta_{M}^{2}$ of those interactions whose relative rank does not change is equal to $0$. Otherwise it will be proportional to the squared Euclidean distance of the effect sizes. When $M$ is small, $\Delta_{M}^{2}$ is usually large, implying that a small $M$ is usually not enough to find all complex interactions and increasing $M$ will lead to large changes in the detected interactions. One way to determine the number of clusters is to inspect values of $\Delta_{M}^{2}$, plotted as a function of $M$, and choose $M$ when $\Delta_{M}^{2}$ approximately converges to $0$. We use this approach to determine $M$ in following experiments (see Figure \ref{fig: choose M} right for an example and Appendix 6 for all plots).

\section{Experiments}
We apply our approach to simulated toy data sets, 3 public real-world data sets with and without injected artificial interactions, and to the MNIST data. We show that our method can identify interpretable interactions with well-characterized uncertainty, and outperforms 3 state-of-the-art baselines.
\label{sec: experiments}
\subsection{Simulated Data}
\label{sec: experiments, sub: sim}
\textbf{Experimental setup:} We simulate datasets with $8$ features and $7$ interaction pairs, using model
\begin{equation}
        y_{i}=\sum_{j=1}^{8}\beta_{j}^{m}x_{j}+\sum_{k=1}^{7}\beta^{i}_{k}h_{k}(x_{k},x_{k+1})+\epsilon
        \label{eq: simulator}
\end{equation}
where $\beta_{j}^{m}$ is the weight of feature $x_j$, $h_{k}(\cdot)$ is the functional form of the $k$th interaction (specified above the panels in Figure \ref{fig: simulated data comparision}) with weight $\beta^{i}_{k}$, and $x_{k},x_{k+1}$ are features involved in this interaction. $x_{j}$ is drawn uniformly from $(0.5, 1.5)$ when $j=1,3,5,6,8$, and from $(-0.5, 0.5)$ when $j=2,4,7$. Noise $\epsilon$ is Gaussian with zero mean and variance adjusted to a specified signal-to-noise ratio. Each simulated dataset includes 20000 samples for training, 5000 for validation, and 5000 for testing.
%
To model interactions, we use a concrete dropout Bayesian neural network with 3 hidden layers of sizes $100$, $100$, and $100$ nodes for $g^{\mathbf{w}}(\mathbf{x})$. During training, we set the length-scale of the Gaussian prior distribution to $10^{-4}$, temperature of the Sigmoid function in the concrete distribution to $0.1$, and the learning rate of Adam to $10^{-3}$.\\
\textbf{Comparison methods:} We apply Bayesian M-GEH and NID on the same trained BNN, using $M=11$ which is determined by $\Delta_{M}^{2}$. NID is applied on the mean NN of the trained BNN, as NID is designed only for deterministic NN.
We implement SHAP interaction score with learning rate equal to $0.01$, and a Lasso regression containing all pairwise multiplicative interactions with regularization set to $5\times10^{-4}$. We include a linear regression model with the correct functional forms for the true interactions and the multiplicative form for all the other interactions as the 'Oracle'. We rank feature pairs according to the \textit{absolute values} of interaction scores from each method from high to low. A good interaction measure should assign the rank of true interactions as close to $1$ as possible. \\
\textbf{Results:} 
In Figure \ref{fig: simulated data comparision}, we increase the signal-to-noise ratio (S/N) gradually from $0$ to see which method can recover the ground-truth interactions with the smallest S/N. The panels in the figure are sorted according to the variance of the corresponding interaction (which is adjusted by weight $\beta^{i}_{k}$). Interactions with higher variances (e.g. $h_{6}$, $h_{4}$) should be easier to detect, because they contribute more to the prediction. We find that \textbf{11-GEH identifies on average the true interactions with a smaller S/N ratio than the other methods}. While it is impossible to assign the highest rank for all interactions, the average rank 11-GEH assigns is by far the highest, and the performance of 11-GEH is close to that of the Oracle. SHAP interaction score works very well for most interactions, but it fails to detect $h_{4}$ and $h_{1}$.
    
\begin{figure}[ht]
    \centering
    \includegraphics[width=1\linewidth]{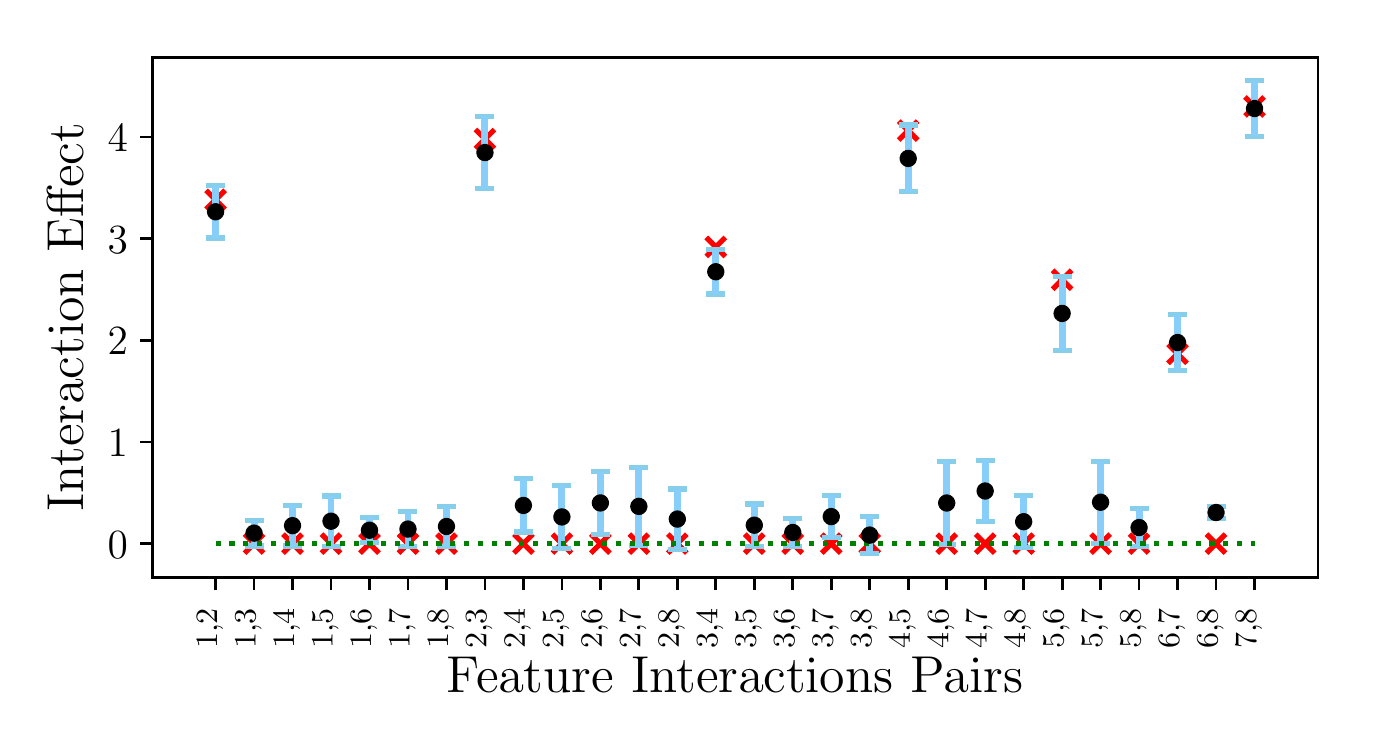}
    \caption{Uncertainty estimation of interaction effects in a simulated dataset. Most of true interaction effects (red crosses) are covered by the corresponding 95\% credible intervals (blue bars).}
    \label{fig: uncertainty}
\end{figure}
\begin{figure}[ht]
    \centering
    \includegraphics[width=1.\linewidth]{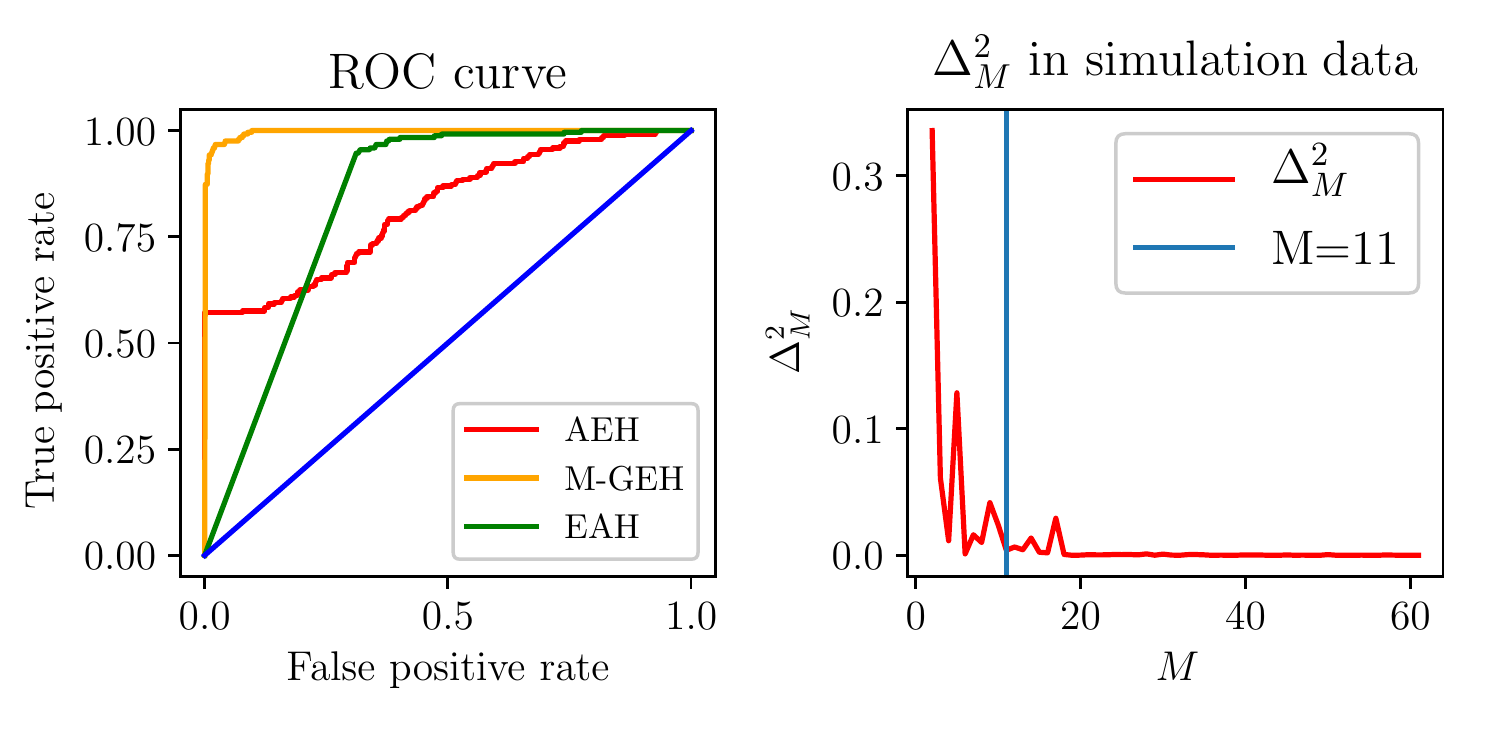}
    \caption{Left: ROC curves for three global interaction measures on 100 datasets with same S/N. Right: Determining the number of clusters based on rank weighted distance. When $M=11$, increasing $M$ will not affect detected interactions significantly.}
    \label{fig: choose M}
\end{figure}

Figure \ref{fig: uncertainty} shows the estimated uncertainty of interaction effects according to Bayesian 11-GEH based on concrete dropout (see Appendix 4 for a mean-field BNN). We observe that \textbf{almost all true interaction effects (red crosses) are covered by the corresponding $95\%$ credible intervals (blue bars)}, centered on the point estimates (black dots). Figure \ref{fig: choose M} left shows the ROC curves for the three global interaction methods introduced above on 100 datasets, where FPR and FNR are calculated based on the decision rule from Section 4.1.3. We see that \textbf{M-GEH has the highest AUC}. Specifically, AEH failed to detect $h_6$ and $h_2$, which are symmetric functions in $\text{dom}(\mathbf{x})$ and hence are cancelled out by AEH (thus a high FNR). On the other hand, EAH failed to reject any false interactions on the basis of its $95\%$ CI (thus a high FPR), so neither method is suitable (see details in Appendix 3). Figure \ref{fig: choose M} right shows $\Delta_{M}^{2}$ as a function of $M$ on simulated data. When $M\geq11$, the detected interactions did not change significantly by increasing $M$ further, hence we selected $M=11$.


\subsection{Public real-world regression datasets}
\label{sec: experiments, sub: real}
\textbf{Datasets:} We analyze 3 publicly available regression datasets: California housing prices, Bike sharing, and Energy efficiency datasets.  California housing prices dataset \cite{pace1997sparse} aims to predict housing prices using 9 features, such as location, number of rooms, number of people, etc. Bike sharing dataset \cite{fanaee2014event} predicts the hourly bike rental count from environmental and seasonal information. Energy efficiency dataset \cite{tsanas2012accurate} aims to predict the load of heating and cooling from the shape of a building. We use $70\%$ of data for training, $20\%$ for validation, and $10\%$ for testing. 
\begin{table}
\centering
\caption{Average FPRs for different M on permuted datasets}
\begin{tabular}{llll}
\hline
\rule{0pt}{12pt}
                   & AEH & M-GEH & EAH \\ \hline
California Housing & 0.015 & 0.097                                                                   & 1.000 \\
Bike Sharing       & 0.017 & 0.042                                                                   & 0.210 \\
Energy Efficiency  & 0.013 & 0.051                                                                   & 0.089\\\hline
\\[-6pt]
\end{tabular}
\label{table: FPRs }
\end{table}
\begin{figure}[ht]
    \centering
    \includegraphics[width=1.\linewidth]{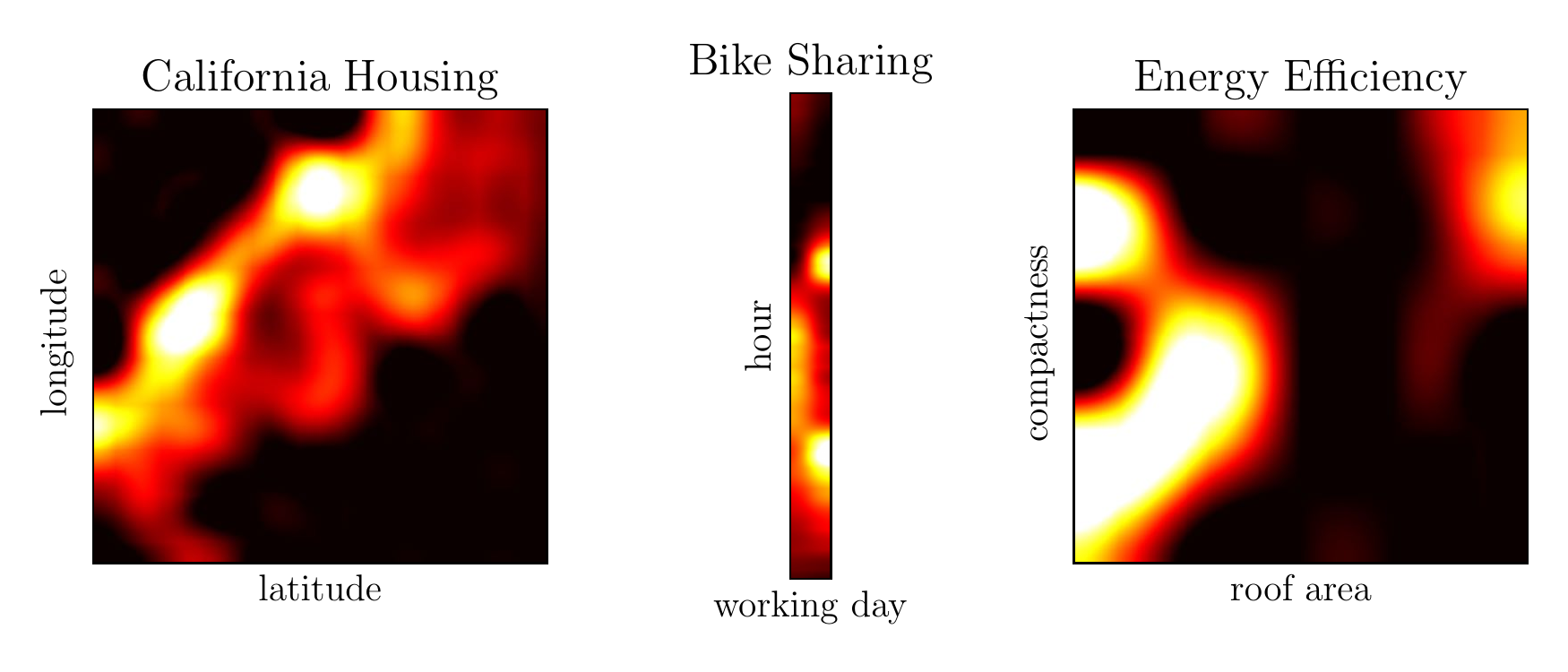}
    \caption{Visualization of one detected interaction for each dataset. The colour shows the value of the regression target (light: high; dark: low).}
    \label{fig: Visualization}
\end{figure}
\noindent\\
\textbf{Experimental setup:} We design $3$ separate experiments to illustrate the benefits of the new proposed method:
\begin{figure*}[ht]
    \centering
    \includegraphics[width=0.8\linewidth]{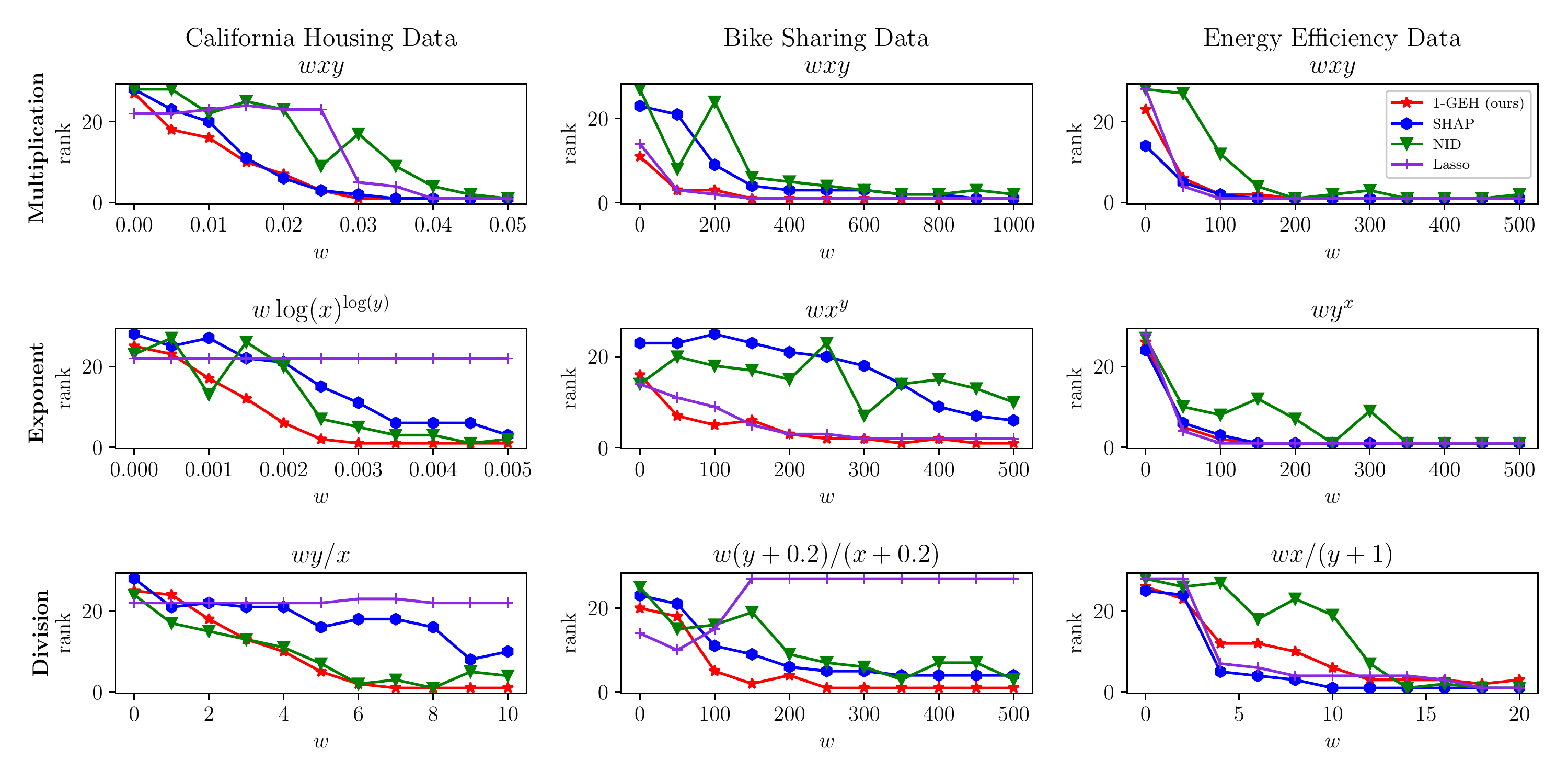}
    \caption{Results on real-world datasets as ranks of the injected interactions for different interaction strengths $w$ (smaller rank is better). Columns correspond to the three datasets, and rows to different types of interactions. The detailed form of the interaction is shown above each panel. In California housing dataset, $x$ and $y$ are the numbers of rooms and households. In bike sharing dataset, $x$ and $y$ are temperature and windspeed. In energy efficiency dataset, $x$ and $y$ are the orientation and glazing area.}
    \label{fig:public}
\end{figure*}
\begin{table*}[t]
\centering
\caption{Top $3$ interactions for real-world datasets without injected interactions}
\begin{tabular}{l l c l c c}
\hline
\rule{0pt}{12pt}
Datasets          & Interacting Features       & M-GEH      & $95\%$ CI        & $P_{\text{Bayes}}$ & $P_{\text{Permute}}$ \\ \hline
                  & total room, population        & $1.532$ & $(0.030, 3.034)$ & 0.026       &   0.000    \\
 California Housing          & longitude, latitude    & $0.901$ & $(0.241, 1.561)$ & 0.003      &   0.000     \\
                  & total room, income & $0.531$ & $(0.065, 0.997)$ & 0.011       &    0.000      \\ \hline
                  & workingday, hour      & $0.337$ & $(0.253, 0.421)$ & 0.001       &  0.000       \\
Bike Sharing        & temperature, humidity       & $0.183$ & $(0.141, 0.225)$ & 0.001       &  0.000        \\
                  & hour, temperature           & $0.180$ & $(0.122, 0.238)$ & 0.002       &  0.000       \\ \hline
                  & roof area, wall area     & $1.223$ & $(0.689, 1.757)$ & 0.000      &    0.000      \\
Energy Efficiency & roof area, height    & $0.938$ & $(0.539, 1.336)$ & 0.001       &      0.000    \\
                  & compactness, roof area          & $0.699$ & $(0.384, 1.013)$ & 0.000       &      0.000    \\    \hline
\\[-6pt]
\end{tabular}
\label{table: result without injection}
\end{table*}
(1) we construct a null hypothesis by permuting the target variable in each dataset, allowing us to estimate FPRs of the different methods, (2) we modify the data sets by injecting artificial interactions to the data sets (without permutation), which establish a ground truth against which we compare the results from the different methods, (3) we analyse the original datasets without injecting any artificial interactions and report and interpret the results.

In detail, in (1) we permute each data set $500$ times. Since all the interactions in permutation datasets are false, if any $95\%$ CI excludes $0$, it will be considered as a false positive, thus we can calculate the FPRs for AEH, M-GEH, and EAH on each dataset. In (2), we inject three types of interactions $I_{in}$: a multiplicative, an exponential, and a division, whose detailed functional forms are given above the panels in Figure \ref{fig:public}. We only inject one interaction at a time. The output is the sum of the true output and injected interactions: $\Tilde{y}=y+I_{in}$. In (3), we explore the results by visualizing some top interactions in each dataset. We also calculate a 'Bayesian' p-value \cite{gelman2013bayesian} for the top interactions using our model and assuming the Gaussian distribution of the interaction score (see Section \ref{sec:geh}), and compare these with the corresponding p-values from permutation, which represents a 'frequentist' score of significance.
%
The same settings are used as in the previous section, except that only one hidden layer with 30 units is used for the third dataset, which has only around $600$ data points. \\
\textbf{Results:} The optimal $M$ for original datasets is 10, 9, and 15 for California housing, bike sharing, and energy efficiency, respectively. 

Table \ref{table: FPRs } shows the results of experiment (1), i.e. estimating the average FPR of different global interaction measures on the $3$  datasets. As expected, AEH has the lowest FPRs, but it is unable to detect complex interactions such as the one between \textit{working day} and \textit{hour} in the bike sharing dataset (see Figure \ref{fig: Visualization}). EAH has the highest FPR, and in the California Housing dataset in particular it considers all false interactions significant. \textbf{The FPRs for M-GEH are approximately correct (close to $5\%$ when using $95\%$ CIs to make decisions)}.

Results with injected interactions (experiment (2)) of varying strengths are shown in Figure \ref{fig:public}. We only use $M=1$ (i.e. AEH) in this experiment to avoid selecting $M$ every time; this gives a lower bound of the performance of out method compared with other methods as it is the weakest interaction measure among the M-GEH family. 
We see that \textbf{1-GEH is consistently the best (or shared best) method in California housing and Bike sharing data sets, irrespective of the type of interaction.} SHAP works well for the simple interaction form (the first row), and small datasets (the third column). We notice that SHAP and Lasso achieve better performance in the third energy efficiency dataset than our method, because it is hard to train a good neural network when the dataset size is small.
\begin{figure*}[ht]
    \centering
    \includegraphics[width=0.8\linewidth]{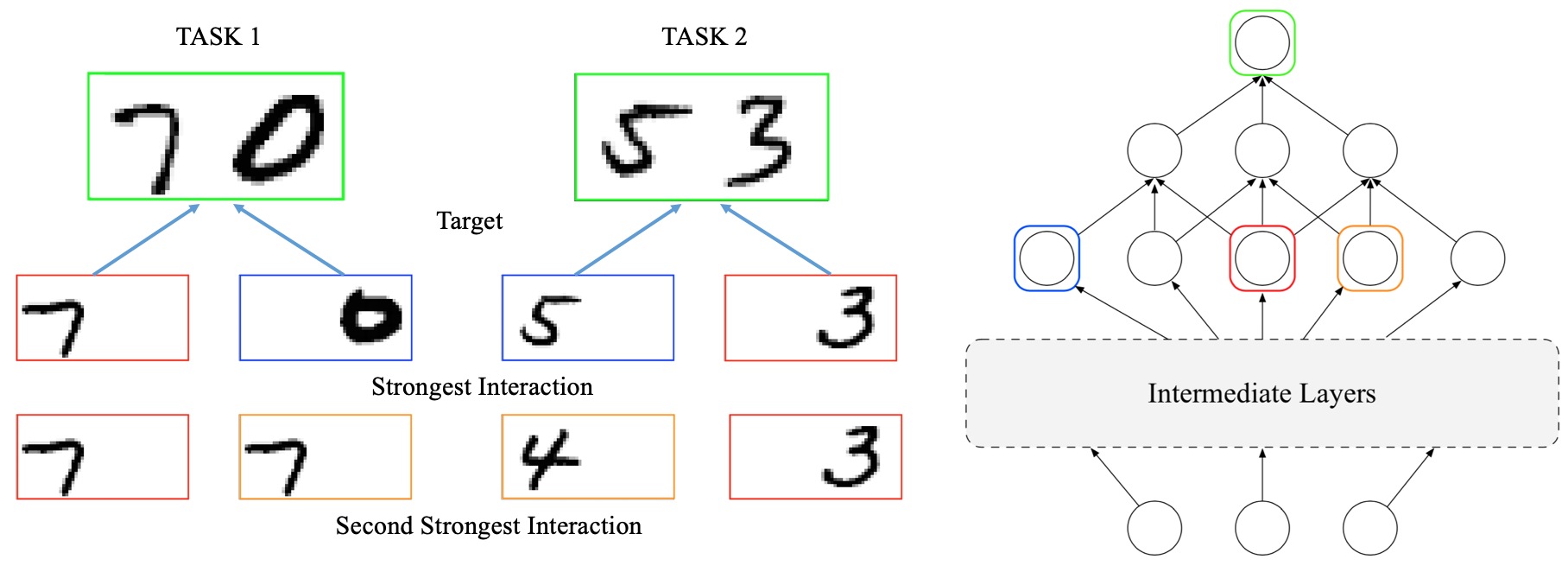}
    \caption{Higher-level feature interactions on MNIST dataset. The structure of the NN is simplified for illustrative purposes.}
    \label{fig: MNIST}
\end{figure*}

Table \ref{table: result without injection} shows results of experiment (3) which contains top $3$ interactions in each dataset (without any injected interactions). M-GEH, CI, and $P_{\text{Bayes}}$ are the estimated means, credible intervals, and Bayesian p-values from our method. We also show a p-value obtained by permuting the target multiple times ($P_{\text{Permute}}$), to create an empirical null distribution of the maximum interaction score (details in Appendix 7). \textbf{All top interactions are meaningful and statistically significant based on both our CIs and permutation}. Examples of strong interactions are shown in Figure \ref{fig: Visualization} (see more in Appendix 5). One strong interaction in the California housing data set is between longitude and latitude, which together specify the location that obviously affects the price. As another example, whether the day is a working day or not will affect the peak hours of bike renting.
\subsection{Higher-level feature interactions in MNIST}
\label{sec: experiments, sub: MNIST}
\textbf{Motivation:} We aim to demonstrate the ability of our method to detect interpretable interactions between higher-level features. Here, higher-level features are those features learned by neural network (i.e. nodes in deeper layers). For this, we design a classification task where the positive label represents a combination of interpretable characteristics of the input. The classification task here is to identify a given combination of two digits, e.g. (5,3), and the inputs are obtained by concatenating randomly chosen MNIST digits. Our expectation is that nodes in upper layers represent interpretable properties of the inputs (e.g. "5 on the left"), such that an interaction between two such nodes corresponds to the positive label (e.g. "5 on the left" and "3 on the right").\\
\textbf{Datasets:} We repeat the experiment twice: the first dataset consists of pairs (7,4), (4,7), (0,4), (4,0), (7,0), (0,7), and the positive label is (7,0); the second dataset consists of pairs (5,4), (4,5), (3,4), (4,3), (5,3), (3,5), and the positive label is (5,3). These pairs and labels are chosen randomly. \\
\textbf{Experimental setup:} We train a LeNet ($2$ convolutional and $3$ fully connected layers) with concrete dropout, and use M-GEH to detect interactions between nodes in the second top fully connected layer, where nodes can be regarded as some high-level features learned by previous layers. Clustering is also implemented on the same layer, and the optimal $M$ for each task is $4$ and $2$, respectively. We provide interpretations for these high-level features by finding one-digit image inputs with white on the other side, e.g. (1,-) or (-,6) that, from all possible one-digit images in the MNIST data, maximize the activation of the node. This is the activation maximization with experts technique for interpreting nodes in intermediate NN layers \cite{erhan2009visualizing}, with empirical distribution of digits in MNIST as the expert.\\
\textbf{Results:} 
Figure \ref{fig: MNIST} shows the top two interactions in the second-highest layer, and presents interpretations for the interacting nodes. We see that all the interacting nodes represent digits related to the prediction tasks instead of unrelated digits, such as 3 in the first task or 7 in the second task. Interestingly, \textbf{in both tasks the strongest interaction is between nodes whose interpretation matches the human intuition exactly}. In the first task, (7,0) is obtained as an interaction between "7 on the left" and "0 on the right", and similarly for the task of classifying (5,3). The second strongest interaction in the (5,3) classification is between nodes with interpretations (4,-) and (-,3), which may be interpreted as excluding digit 4 on the left, when there is 3 on the right. The interaction between nodes which both have interpretation (7,-) may be related to learning different parts of digit 7.
\section{Conclusion}
\label{sec: conclusion}
We presented a novel method to learn global pairwise interactions using Bayesian neural networks. Our work addresses the problem of providing second-order explanations for neural network predictions, which, unlike the case of first-order explanations, is relatively understudied in the field of interpretable machine learning. In addition, the method provides a practical means to extract meaningful information about complex feature interactions from real-world data sets. Based on the Hessian, the method is intuitive and easy to calculate, and, unlike the alternatives, does not require further investigation of the structure of the neural network. Other strengths of the method are that it provides uncertainty for the detected interactions, and comes with appealing theoretical properties which ensure that by improving the underlying BNN, interaction detection can be improved. The method empirically outperformed several state-of-the-art baselines, and, for the first time, we also demonstrated the ability to learn interpretable interactions between higher-level features. 


To estimate global effects for complex interactions, it is necessary to merge information from possibly conflicting regions in the input space. We resolved this by clustering the data in the input space and aggregating interaction scores across the clusters. A simple $L_2$ distance was used for clustering. Combined with a heuristic to define the number of clusters, this approach performed consistently well in the experiments. However, this may not the  optimal choice as it ignores interactions when calculating the distance. We leave it for future work to explore alternative ways to define the clustering.

\section{Appendix}
\label{sec: Appendix}
Please find the appendix on:\\
https://github.com/aalto-ml4h/InteractionDetection


\end{document}